# Improving predictions by nonlinear regression models from outlying input data

William W. Hsieh[1]

Department of Earth, Ocean and Atmospheric Sciences,
University of British Columbia, Vancouver, BC, Canada

## Abstract

When applying machine learning/statistical methods to the environmental sciences, nonlinear regression (NLR) models often perform only slightly better and occasionally worse than linear regression (LR). The proposed reason for this conundrum is that NLR models can give predictions much worse than LR when given input data which lie outside the domain used in model training. Continuous unbounded variables are widely used in environmental sciences, whence not uncommon for new input data to lie far outside the training domain. For six environmental datasets, inputs in the test data were classified as "outliers" and "non-outliers" based on the Mahalanobis distance from the training input data. The prediction scores (mean absolute error, Spearman correlation) showed NLR to outperform LR for the non-outliers, but often underperform LR for the outliers. An approach based on Occam's Razor (OR) was proposed, where linear extrapolation was used instead of nonlinear extrapolation for the outliers. The linear extrapolation to the outlier domain was based on the NLR model within the non-outlier domain. This $NLR_{OR}$ approach reduced occurrences of very poor extrapolation by NLR, and it tended to outperform NLR and LR for the outliers. In conclusion, input test data should be screened for outliers. For outliers, the unreliable NLR predictions can be replaced by $NLR_{OR}$ or LR predictions, or by issuing a "no reliable prediction" warning.

## 1 Introduction

The widespread application of machine learning (ML) to commerce, engineering and a myriad of other fields in recent decades has encouraged environmental scientists to use ML models. However, unlike the great successes attained in the non-environmental fields, the achievements of ML in the environmental sciences have been on the whole much more modest. Often the results from properly regularized nonlinear ML models are only slightly better than traditional linear statistical model results and occasionally worse (Yuval and Hsieh, 2002; Zeng et al., 2011a, 2011b; Peng et al., 2017; Mao and Monahan, 2018) – this conundrum has caused many environmental scientists to be unwilling to use ML methods, as ML methods also have the reputation of being "black boxes" due to their poor interpretability when compared to the more transparent linear statistical models.

Why are the ML results not as impressive in the environmental sciences as in the non-environmental fields? For climate applications, the process of averaging daily data to form climate data tends to linearize the relation between the input and output variables according to the central limit theorem, thereby diminishing the advantage of the nonlinear ML models (Yuval and Hsieh, 2002). However, even in the absence of the central limit theorem effect, nonlinear ML results are still often not much better than linear regression (LR) results (Peng et al., 2017; Mao and Monahan, 2018), so there needs to be at least another reason for the underperformance.

---

[1] Current address:
4028 Hopesmore Dr., Victoria, BC, V8N 5S9, Canada
Tel.: +1-250-388-0508,  E-mail: whsieh@eos.ubc.ca

There is one major difference between non-environmental and environmental datasets. Most non-environmental datasets contain discrete input data (usually with a finite domain, e.g. colour pixels where the intensities of the red, green and blue components are integers from 0 to 255) and categorical input data (e.g. alphabets and numerical digits in texts). In contrast, most environmental datasets contain continuous input variables (e.g. temperature, air pressure, wind speed, precipitation amount, pollutant concentration, etc.). To predict output variables, one would use a classification model if the output is discrete or categorical and a regression model if the output is continuous. Again, classification is much more common in non-environmental datasets, and many ML methods were developed first for classification and later modified for regression, e.g. support vector machines (Cortes and Vapnik, 1995; Vapnik et al., 1997).

When given input values lying outside the training domain, most nonlinear ML models (e.g. artificial neural networks, support vector machines, Gaussian processes) would perform nonlinear extrapolation (Hsieh, 2009, pp.303-305). A notable exception is the random forest (Breiman, 2001): as the input extends beyond the boundary of the training domain, instead of linear or nonlinear extrapolation, the model gives a somewhat unrealistic constant output beyond the boundary of the training domain.

When applying the back-propagating artificial neural network (ANN) classifier to fault diagnosis, Kramer and Leonard (1990) found that ANN did not perform well compared to the simple 1-nearest neighbour classifier when there was extrapolation. The problem arose from the ANN assigning arbitrary classifications to regions of the input space which were devoid of training data. Leonard and Kramer (1991) made comparisons between ANN and radial basis function classifier on their extrapolation performance. Barnard and Wessels (1992) made further distinctions between interpolation and extrapolation, with the ANN classifier performing well with interpolation but not so well when there is extrapolation. The extrapolation problem also gets worse with higher dimensional input space due to the "curse of dimensionality" (Bellman, 1961; Barnard and Wessels, 1992).

In contrast to discrete data (usually with finite upper and lower bounds) or categorical data, continuous data are usually not bounded, so new continuous input data can lie far outside the domain of input variables used in training the model. Thus the extrapolation problem can be much worse when applying ML methods to enviromental science problems, which tend to use unbounded continuous input data.

The objectives of this paper are to investigate (i) whether the presence of such "outliers" in the inputs and the resulting nonlinear extrapolation would lead to the nonlinear regression (NLR) models underperforming the LR models for these outliers, and (ii) whether avoiding nonlinear extrapolation at the outliers could improve the ML model predictions. In pursuing objective (ii), one is tempted to follow the wisdom of Occam's Razor, i.e. a preference for the simple linear model over the complicated nonlinear model for the outliers. Six different environmental datasets (Appendix A) were used to test models in this paper.

## 2 Nonlinear regression

The feed-forward artificial neural network (ANN) model with one layer of hidden nodes is probably the most commonly used ML model (Bishop, 1995, 2006; Hsieh, 2009). With input $\mathbf{x} \in \mathbb{R}^d$ and output $\hat{\mathbf{y}} \in \mathbb{R}^m$, the model has output

$$\hat{y}_k = \sum_{i=1}^{L} \beta_{ki} h(\mathbf{w}_i \cdot \mathbf{x} + b_i) + \beta_{k0}, \quad (k=1,...,m), \qquad (1)$$

where $L$ is the number of hidden nodes, $\mathbf{w}_i$ and $b_i$ are, respectively, the weight and bias parameters in the hidden layer, $\beta_{ki}$ and $\beta_{k0}$ are, respectively, the weight and bias parameters in the output layer and $h$ is the activation function at the hidden layer. In this paper, we will try three different activation functions: the sigmoidal function

$$h(x) = \frac{1}{1+\exp(-x)}, \qquad (2)$$

the Gaussian or radial basis function

$$h(x) = \exp(-x^2), \qquad (3)$$

and the softplus function (Glorot et al., 2011)

$$h(x) = \log(1+e^x). \qquad (4)$$

The various weight and bias parameters are solved by minimizing the mean squared error (MSE) between the output $\hat{\mathbf{y}}$ and the target data $\mathbf{y}$.

There are many variants of the ANN model – we will use a randomized neural network, also called an extreme learning machine (ELM), which is structurally a feed-forward ANN model with one layer of hidden nodes. However, in ELM, the weights in the hidden layer (i.e. $\mathbf{w}_i$ and $b_i$) are chosen randomly, hence only the weights in the output layer ($\beta_{ki}$ and $\beta_{k0}$) require optimization (Schmidt et al., 1992; Huang et al., 2011), and even the $\beta_{k0}$ term can be omitted (Huang, 2014).

Thus, in the ELM approach, training the feed-forward neural network with $N$ samples simply involves finding the least-squares solution of the linear system $\mathbf{HB} = \mathbf{Y}$ (Huang et al., 2011), where the hidden layer output matrix $\mathbf{H}$ of dimension $N \times L$ is

$$\mathbf{H} = \begin{bmatrix} h(\mathbf{w}_1 \cdot \mathbf{x}_1 + b_1) & \cdots & h(\mathbf{w}_L \cdot \mathbf{x}_1 + b_L) \\ \vdots & \ddots & \vdots \\ h(\mathbf{w}_1 \cdot \mathbf{x}_N + b_1) & \cdots & h(\mathbf{w}_L \cdot \mathbf{x}_N + b_L) \end{bmatrix}, \qquad (5)$$

and the $\mathbf{B}$ weight matrix of dimension $L \times m$ and the target data matrix $\mathbf{Y}$ of dimension $N \times m$ are

$$\mathbf{B} = \begin{bmatrix} \boldsymbol{\beta}_1^T \\ \vdots \\ \boldsymbol{\beta}_L^T \end{bmatrix}, \text{ and } \mathbf{Y} = \begin{bmatrix} \mathbf{y}_1^T \\ \vdots \\ \mathbf{y}_N^T \end{bmatrix}, \qquad (6)$$

with superscript T denoting the transpose and $\boldsymbol{\beta}_i^T = [\beta_{i1}, ..., \beta_{im}]$.

The linear system for $\mathbf{B}$ is solved via the Moore-Penrose generalized inverse $\mathbf{H}^\dagger$ (Huang et al., 2006), i.e.

$$\hat{\mathbf{B}} = \mathbf{H}^\dagger \mathbf{Y}, \text{ with } \mathbf{H}^\dagger = (\mathbf{H}^T\mathbf{H})^{-1}\mathbf{H}^T. \qquad (7)$$

Instead of nonlinear optimization with traditional ANN, ELM requires only linear least squares optimization (as used in LR), which generally allows ELM to run thousands of times faster than ANN (Huang et al., 2006). Lima et al. (2015) tested the ELM, ANN, support vector regression and random forest on nine different environmental datasets for regression, and found that ANN and ELM had similar skills but ELM was much faster than ANN except for large datasets.

If a feed-forward neural network has $d$ inputs, one layer of $L$ hidden nodes and $m$ outputs, the number of adjustable parameters (i.e. weights) is $Lm$ for ELM versus $(d+1)L+(L+1)m$ for the traditional ANN. In this paper, only datasets with $m = 1$ were used, so the number of adjustable weights in ELM equals the number of hidden nodes. The optimal value for $L$ was determined by a 5-fold cross validation (Bishop, 2006; Hsieh, 2009) of the ELM on the training data.

This is the batch version of the ELM. For online learning, i.e. where the model is required to be updated continually with new data, there is online sequential ELM, which updates the ELM model efficiently using a recursive linear least squares algorithm (Liang et al., 2006; Lima et al., 2016), and a further extension which allows the number of hidden nodes to self-adapt (i.e. increase or decrease) as online learning proceeds (Lima et al., 2017).

In this paper the ELM batch model, based on the Matlab code from Huang et al. (2006), was run 100 times with different random weights, giving an ensemble of 100 members. The NLR model output was the averaged output of the 100 ELM ensemble members.

## 3  Nonlinear extrapolation

First, a toy problem was used to illustrate the effect of extrapolation when the input variable lies outside the domain of the training data. The "true" signal was chosen to be $y = x + 0.2\,x^2$, where $x$ was a Gaussian random variable with zero mean and unit standard deviation. Gaussian noise with twice the standard deviation of the $y$ signal was added to generate the training data. LR and NLR models were trained using a training dataset of 100 points. Fig. 1 shows the results from using three different activation functions.

Within the domain of the training data, there is very little difference between the true signal, the individual ensemble members and the ensemble average of the NLR ensemble in Figs. 1a, b and c. However, outside of the training domain, there is great divergence among the true signal, the individual members and the ensemble average. The NLR (ensemble averaged) solutions using the three different activation functions are very similar within the training domain but are quite different outside (Figs. 1a, b and c). Within the training domain, the NLR solution is closer to the true signal than LR, however, that may not be the case when outside the training domain (e.g. when $x > 3$ in Fig. 1b).

By following the gradient of the NLR function at the boundary of the training domain, linear extrapolation of the NLR solution beyond the training domain was performed. The linear extrapolation of the NLR model is closer to the true solution than the NLR model in Figs. 1a, b and c.

The three activation functions have very different asymptotic behaviour (Fig. 1d). As the sigmoidal activation function is bounded as $x \rightarrow \pm\infty$ in Eq. (2), the neural network output, being a linear combination of the activation functions in Eq. (1), is also bounded asymptotically. The radial basis function in Eq. (3) asymptotes to zero, so the model output also asymptotes to zero. The softplus function in Eq. (4) asymptotes linearly as $x \rightarrow +\infty$, hence the model output asymptotes linearly. Thus the particular choice of an activation function does affect the extrapolation of the NLR model to beyond the training domain as noted by Hsieh (2009, pp.303-305).

## 4  Extrapolation in environmental datasets

The six environmental variables (see Appendix A for details) consist of four variables from British Columbia, Canada – daily streamflow at Englishman River (ENG) and at Stave River (STA), daily precipitation at Vancouver International Airport (YVR), daily suspended sediment concentration in Fraser River (FRA), hourly wind speed at Jersey Airport (JER), and hourly particulate matter $PM_{2.5}$ concentration in Beijing, China (BEI). The two streamflow variables have very different behaviour – ENG is primarily driven by rainfall, with peak streamflow during the wet winter season but STA is driven mainly by snow melt, with peak streamflow during the dry early summer period (Lima et al., 2017, Fig.1). From the six environmental datasets, we used the earlier part of the data record for training and the later part for testing. We also reversed the order, using the later part for training and the earlier part for testing (ENGr, STAr, YVRr, FRAr, JERr and BEIr) (Table 1).

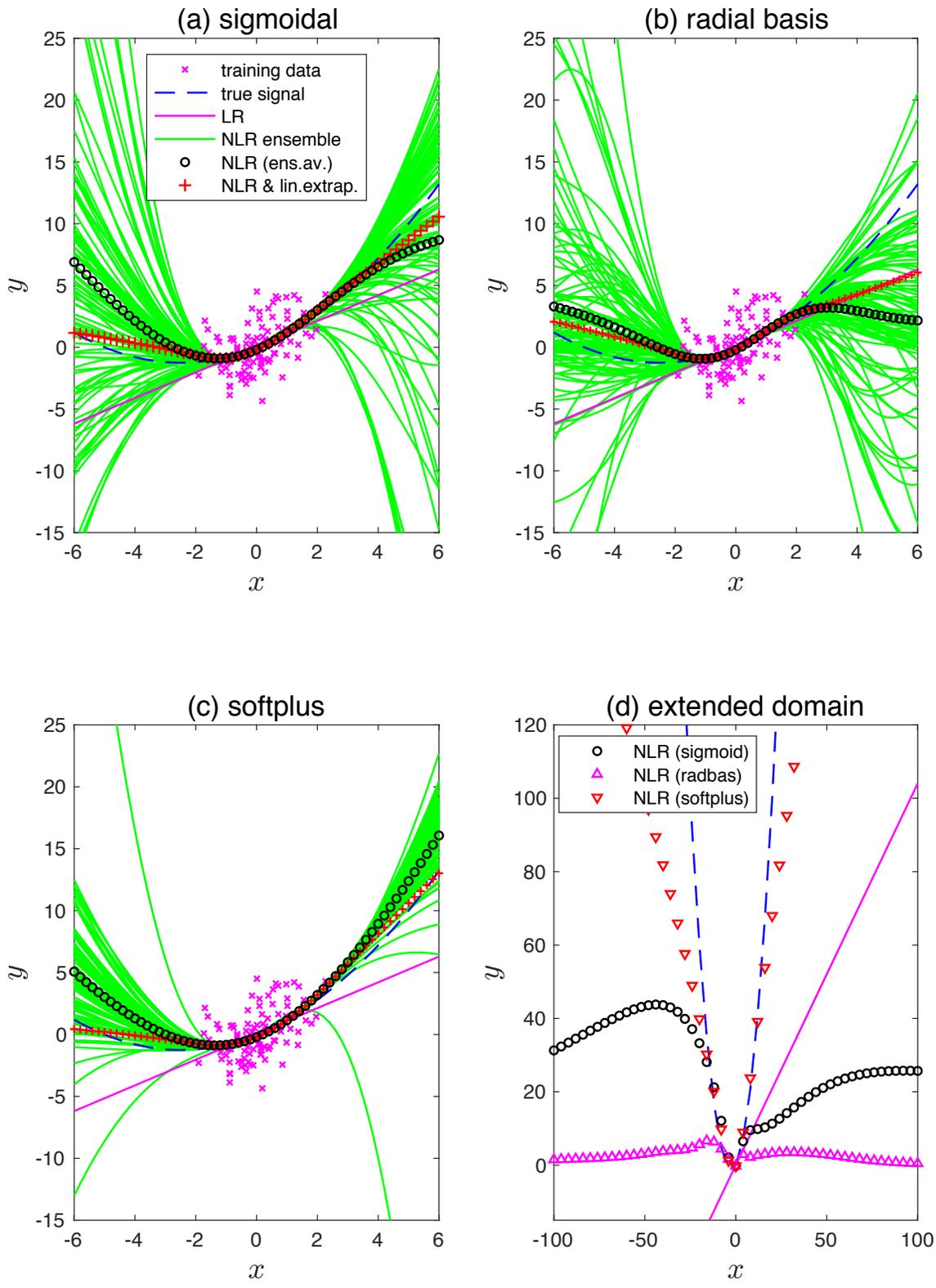

Figure 1: A toy problem illustrating the effect of extrapolation with the (a) sigmoidal, (b) radial basis and (c) softplus activation function used in the ELM NLR model. The training data, true signal, linear regression, the NLR ensemble with 100 members, the NLR ensemble averaged output, and linear extrapolation used to extend beyond the outermost training data for the NLR model are shown. (d) shows extrapolation over an extended domain for the three activation functions, as well as the true signal (dashed) and LR.

Table 1: Properties of the Environmental Datasets: station name (with "r" indicating later part of record used for training), number of data points and period used for training and testing, $r_{outl}$ (a measure of the severity of outliers as defined in Eq. (10)), number of outliers in the test data with outliers defined as exceeding the 99th and 95th percentile of the Mahalanobis distance of the training data, and number of hidden nodes used with the sigmoidal, radial basis and softplus activation functions, as determined by a 5-fold cross-validation on the training data. The percentage of outliers beyond the 95th percentile to the total number of testing data is given in parenthesis.

| Stn | Training | | Testing | | $r_{outl}$ | # outliers | | # hidden nodes | | |
|---|---|---|---|---|---|---|---|---|---|---|
| | # pts | period | # pts | period | | 99% | 95% | sigm | radb | softp |
| ENG | 1807 | 1985-1989 | 7733 | 1990-2011 | 4.00 | 87 | 424(5.5) | 150 | 150 | 100 |
| ENGr | 1807 | 2007-2011 | 7733 | 1985-2006 | 2.31 | 58 | 406(5.3) | 150 | 140 | 150 |
| STA | 1786 | 1985-1989 | 7834 | 1990-2011 | 6.22 | 991 | 1837(23.4) | 300 | 300 | 250 |
| STAr | 1786 | 2007-2011 | 7834 | 1985-2006 | 3.99 | 196 | 638(8.1) | 270 | 330 | 220 |
| YVR | 977 | 1971-1976 | 4003 | 1977-2000 | 1.38 | 57 | 217(5.4) | 23 | 23 | 21 |
| YVRr | 977 | 1995-2000 | 4003 | 1971-1995 | 1.27 | 38 | 126(3.1) | 20 | 21 | 21 |
| FRAr | 1095 | 1977-1979 | 2557 | 1970-1976 | 2.04 | 157 | 423(16.5) | 19 | 27 | 21 |
| JER | 899 | 1980-1989 | 2063 | 1990-2012 | 1.24 | 24 | 86(4.2) | 43 | 50 | 34 |
| JERr | 895 | 2003-2012 | 2067 | 1980-2002 | 1.26 | 17 | 106(5.1) | 25 | 30 | 26 |
| BEI | 24417 | 2010-2012 | 25162 | 2013-2015 | 1.43 | 152 | 957(3.8) | 60 | 75 | 45 |
| BEIr | 25162 | 2013-2015 | 24417 | 2010-2012 | 7.73 | 235 | 963(3.9) | 115 | 150 | 125 |

To determine which test input data were "outliers", the Mahalanobis distance was used (Mahalanobis, 1936). For the $d$ input variables, the training dataset allowed us to compute the sample mean vector $\boldsymbol{\mu} = [\mu_1,\ldots, \mu_d]$ and the sample covariance matrix $\mathbf{C}$. The Mahalanobis distance for an input $\mathbf{x}$ is

$$D_{\mathrm{M}}(\mathbf{x}) = \sqrt{(\mathbf{x}-\boldsymbol{\mu})^{\mathrm{T}} \mathbf{C}^{-1} (\mathbf{x}-\boldsymbol{\mu})}. \tag{8}$$

$D_M$ was computed (using the Matlab function mahal) for the training input data, and the 99th percentile of the Mahalanobis distance was considered the boundary for outliers. For each test input data point, its nearest neighbour from the training set was also determined based on the Euclidean distance. A test input data point having (i) a Mahalanobis distance (computed using Eq. (8) with $\boldsymbol{\mu}$ and $\mathbf{C}$ from the training dataset) exceeding the 99th percentile boundary and (ii) a (Euclidean) distance to the centre (i.e. median) of the training data exceeding the distance between its nearest neighbour and the centre was considered an outlier. The condition (ii) is needed to prevent a small number of test data which satisfy (i) from being labelled "outliers" as they are actually closer to the centre of the training data than their nearest neighbours are to the centre.

Fig. 2 shows the mean absolute error normalized (by the median absolute deviation) (MAEn) for the "outliers" and "non-outliers" in the testing data. The station FRA had only 3 outliers, so it was omitted from Fig. 2. The NLR model (being the ensemble average of 100 members) was repeated for 200 trials with different initialization of the random number generator. Of the three activation functions used, radial basis occasionally had an ensemble member performing much worse than the rest, so the ensemble averaging for the radial basis runs omitted the maximum value and the minimum value in the 100 ensemble values.

For the non-outliers in the test data, NLR has lower MAEn than LR, with the exception of ENG and ENGr using radial basis (Fig. 2b). However, the advantage of NLR over LR largely disappeared for the outliers in the test data (Fig. 2a). Also, some of the trials have MAEn lying well above the upper whisker in the boxplot, especially for the particulate matter concentration ($PM_{2.5}$) at Beijing (BEIr) and the streamflow at STA, indicating very poor results from the nonlinear extrapolation for some trials.

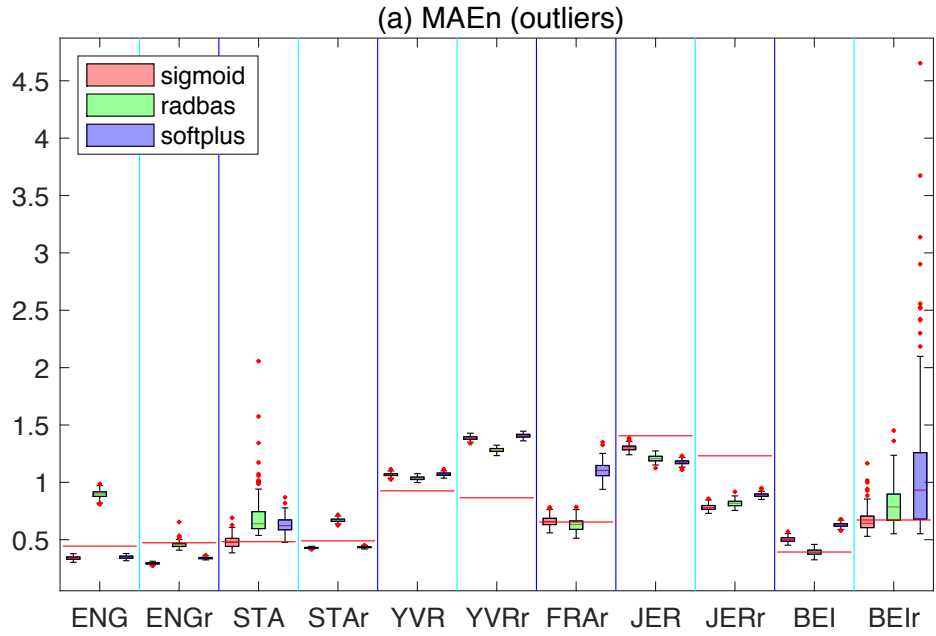

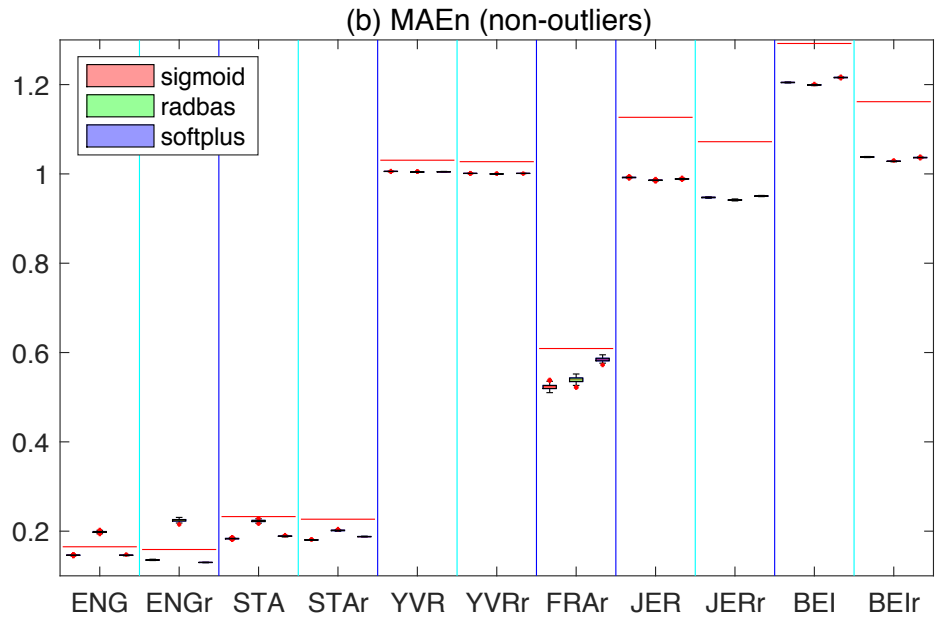

Figure 2: Boxplot of the mean absolute error (normalized by the median absolute deviation) for the (a) outliers and (b) non-outliers in the test data, with three types of activation functions used in the NLR model. The "waistline" of a box marks the median MAEn of the 200 trials, while the bottom and top of each box mark the 25th and 75th percentiles, respectively. The distance between these two percentiles is the interquartile range (IQR), and the upper/lower whisker extends from the top/bottom of the box to the highest/lowest value within 1.5 IQR, with data beyond the whiskers plotted as small crosses. The corresponding MAEn for LR are shown as horizontal lines.

For the correlation score between the model output and the target data in Fig. 3, the more robust Spearman correlation (Spearman, 1904) was used instead of the Pearson correlation (where Spearman correlation involves computing the Pearson correlation of the ranked variables). Again the tendency of NLR to outperform LR for the non-outliers (Fig. 3b) disappeared for the outliers (Fig. 3a).

The difference between MAEn from NLR and MAEn from LR is shown in Fig. 4. With the "waistline" of a box marking the median MAEn of the 200 trials, the dash and dot-dash lines in Fig. 4 indicate, respectively, the mean and the median position of the waistline over all the datasets. In Fig. 4b, both the dash and dot-dash lines lie below the horizontal axis, indicating NLR generally outperforming LR in MAEn for non-outliers. However, in Fig. 4a, the two lines have shifted upward for the outliers, so the dash line lies above the horizontal axis indicating NLR generally underperforming LR, while the dot-dash line is hardly visible as it overlaps with the horizontal axis. Again, some of the trials have MAEn(NLR) − MAEn(LR) lying well above the upper whisker in the boxplot, especially for BEIr and STA, indicating very poor results from the nonlinear extrapolation.

The difference between the Spearman correlation from NLR and that from LR in Fig. 5 again has the dash and dot-dash lines lying above the horizontal axis in (b) indicating NLR generally outperforming LR for non-outliers, and lying below below the horizontal axis in (a) indicating NLR generally underperforming LR for outliers.

From our environmental datasets, we have learned that while NLR generally outperforms LR for non-outlying input test data, NLR often performs worse than LR for outliers. The next question is: can we avoid the ill effects of nonlinear extrapolation?

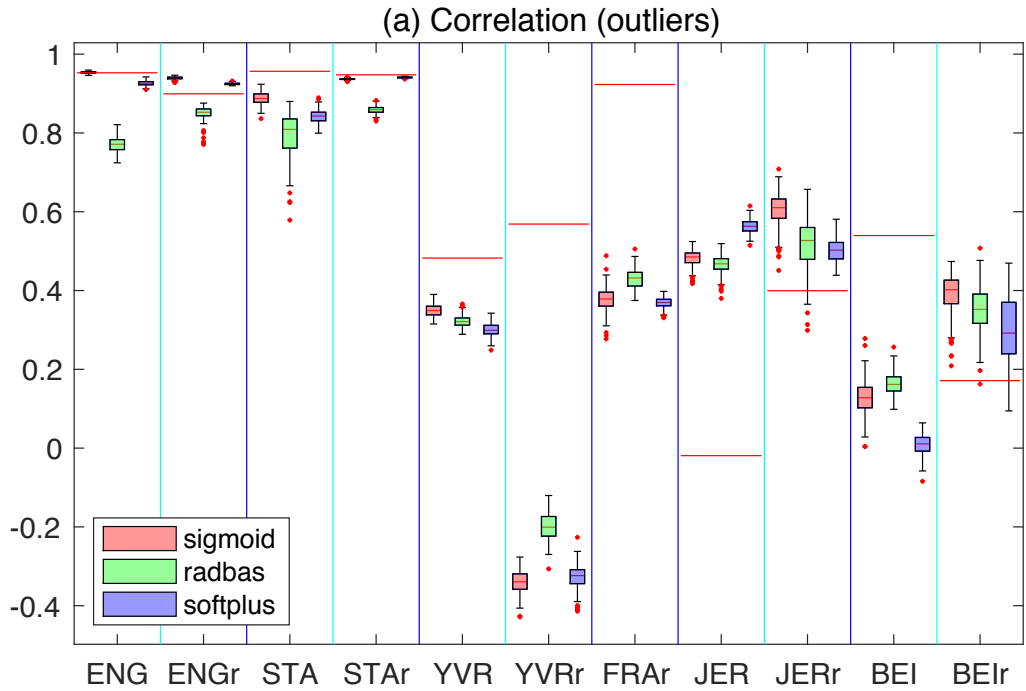

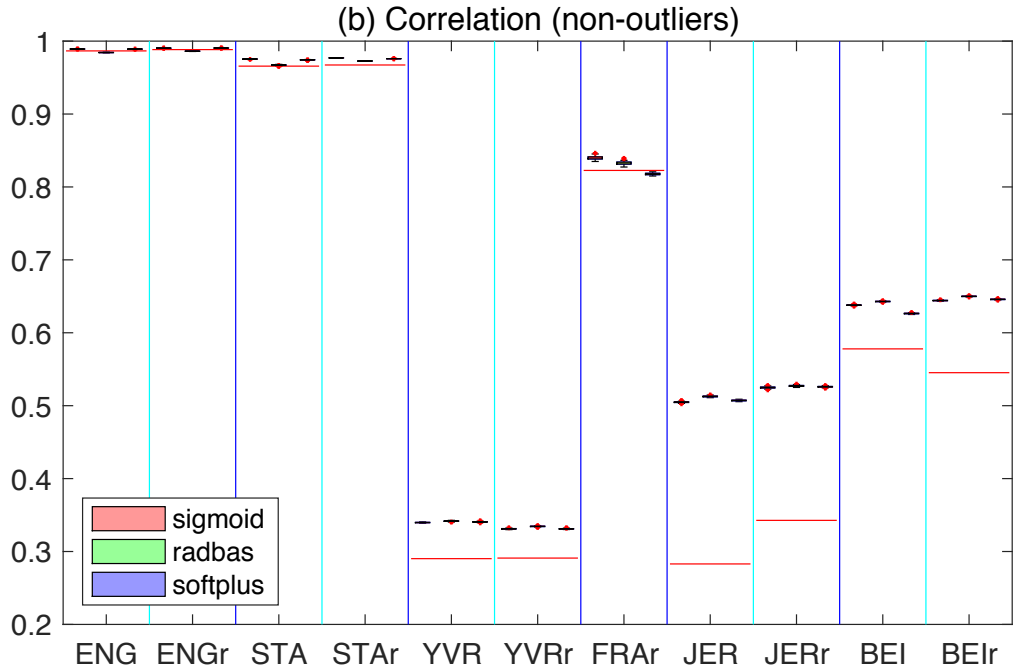

Figure 3: Boxplot of the Spearman correlation for the (a) outliers and (b) non-outliers in the test data, with three types of activation functions used in the NLR model, and with the corresponding LR results shown as horizontal lines.

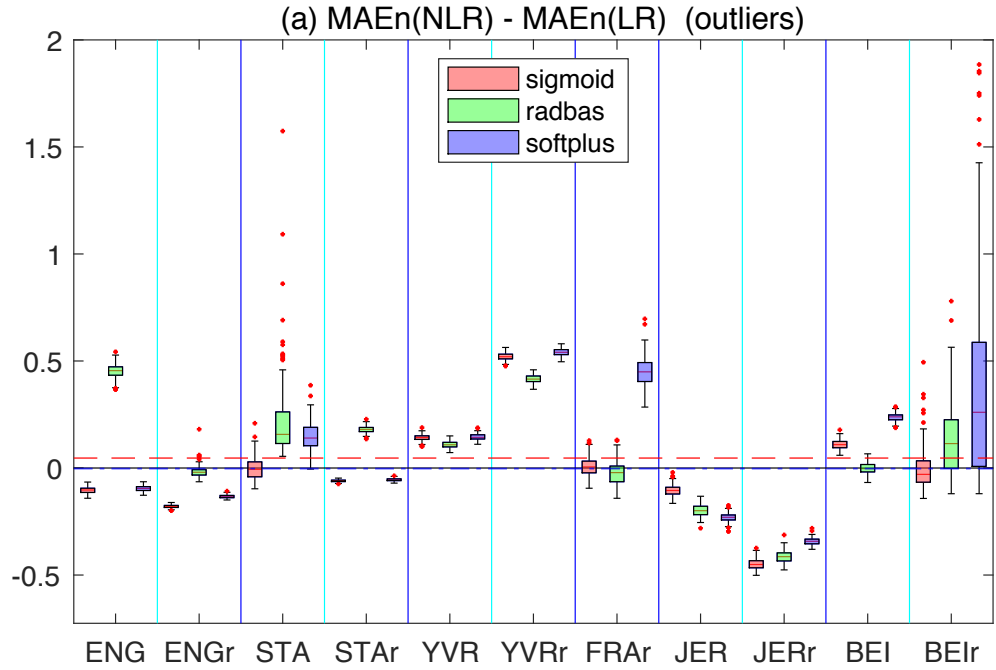

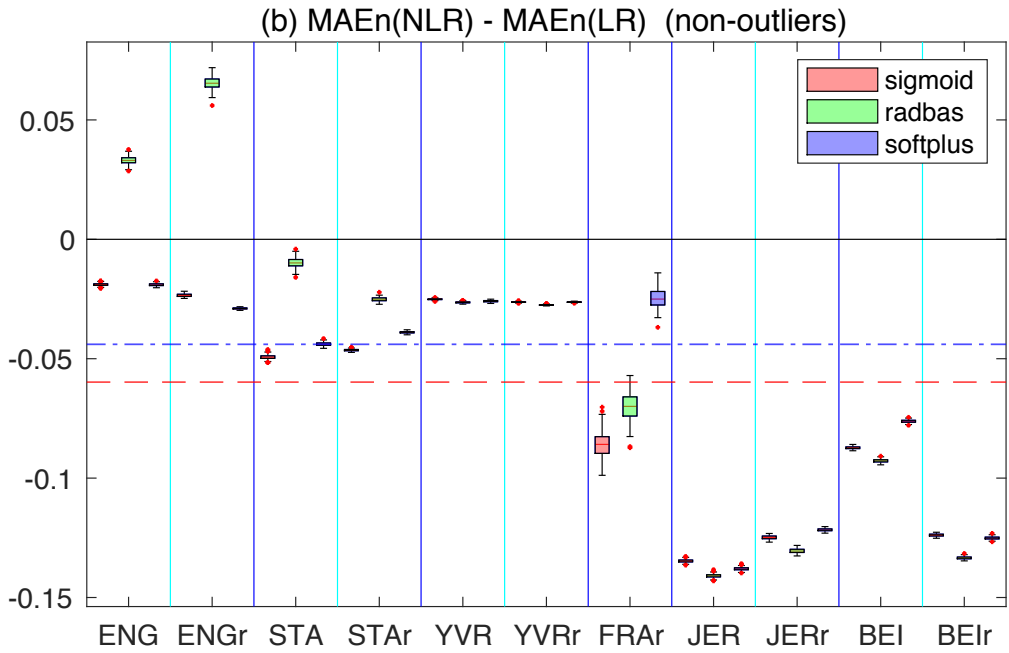

Figure 4: Boxplot of the difference in the normalized MAE between the NLR and LR models for the (a) outliers and (b) non-outliers in the test data, with three types of activation functions used in the NLR model. The mean and the median position of the waistline over all the datasets are indicated by the horizontal dash and dot-dash lines, respectively. The rightmost box in (a) has four distant small crosses in the range (2,4) which are not shown.

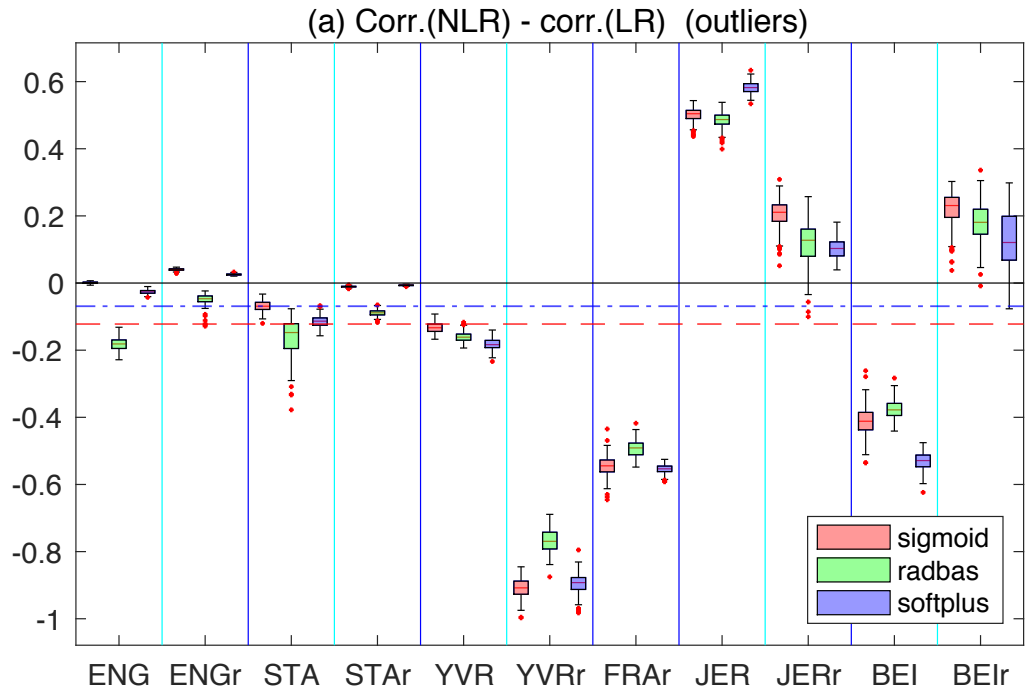

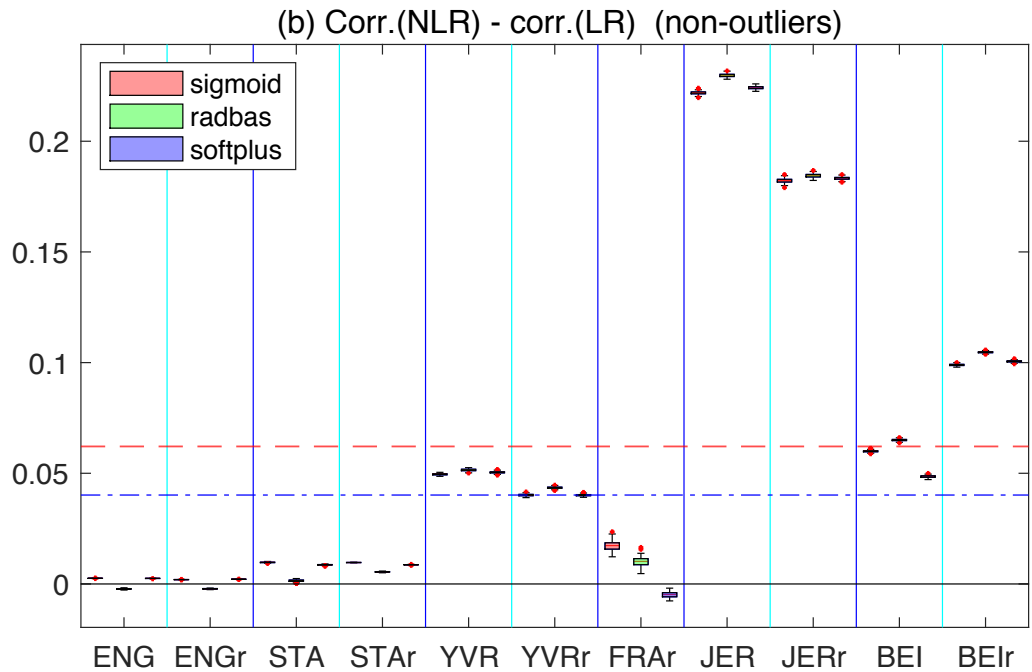

Figure 5: Boxplot of the difference in the Spearman correlation between the NLR and LR models for the (a) outliers and (b) non-outliers in the test data, with the mean and the median position of the waistline over all the datasets indicated by the horizontal dash and dot-dash lines, respectively.

# 5 Occam's razor in extrapolation

Occam's razor is the principle of parsimony, attributed to the English philosopher and Franciscan friar William of Occam (c. 1285–1349), which states that of two competing theories, the simpler explanation of an entity is to be preferred. For our problem, in the non-outlying domain, the training data provide support for the NLR, but over the outlying domain, where there are no training data, Occam's razor would prefer a simple linear model over a complicated nonlinear model. In Figs. 4a and 5a, simply using LR instead of NLR on outliers tended to improve prediction performance. Alternatively, in our toy model with only a single input variable (Fig. 1), we have proposed performing linear extrapolation over the outlying domain based on the gradient of the NLR model at the furthest extent of the training data domain. We now extend this approach to higher dimensional input space.

For the dataset YVR, with 3 inputs variables ($x_1$, $x_2$, $x_3$), Fig. 6 shows the training and testing data in the $x_1$-$x_3$ space, with the outlying test data circled. Since the dimension $x_2$ was not plotted, some test data outlying in the $x_2$ dimension appeared to lie well within the training data cluster. For the leftmost outlying test data point, linear extrapolation from two directions is illustrated: from the nearest neighbour (Fig. 6a) and along the direction of the centre of training data (Fig. 6b).

Linear extrapolation requires estimating the gradient of the NLR function from two points. In Fig. 6a, the two points are the nearest neighbour and another point further along the line connecting the outlier and the nearest neighbour (as marked by the star). The star point is a distance $\delta_1 d_{nn}$ from the nearest neighbour and $(1+\delta_1) d_{nn}$ from the outlier, where $d_{nn}$ is the distance between the outlier and the nearest neighbour and $\delta_1$ is a chosen positive scalar.

In Fig. 6b, the two points used for estimating the gradient of the NLR function are the solid circle (projection of the nearest neighbour onto the line connecting the outlier and the centre of the training data) and the point marked by a star which lies along that line. If $d_c$ is the distance between the solid circle and the centre, then the star is a distance $\delta_2 d_c$ from the solid circle, with $\delta_2$ a chosen positive scalar.

Mathematically, one would want to use a small $\delta_1$, i.e. having the star point close to the nearest neighbour in Fig. 6a, for a more accurate estimate of the gradient. However, the accuracy of the NLR function drops as it approaches the boundary of the training domain due to sparsity of data. Similarly, in Fig. 6b, having the star point further from the solid circle and closer to the centre allows it to use a more accurate part of the NLR model.

One can perform linear extrapolation for outliers multiple times using different values of $\delta_1$ and $\delta_2$ and take the mean or median. The scheme used here is to linearly extrapolate (a) from nearest neighbour twice using $\delta_1 = 0.25$ and $0.5$, and (b) from the direction of the centre of the training data twice using $\delta_2 = 0.5$ and $1$ (where $\delta_2 = 1$ has the star point located right at the centre). The final solution for outliers was taken to be the median of five values – the four linear extrapolation values and the NLR value. The NLR value was included in the median in case of inconsistent linear extrapolations, e.g. the extrapolation by (a) might yield high values while (b) might yield low values relative to the NLR value. This resulting model from following Occam's Razor will henceforth be referred to as the NLR$_{OR}$ model.

A slight modification of NLR$_{OR}$ was needed for the Beijing PM$_{2.5}$ dataset (BEI and BEIr), as it contained categorical input indicating four possible wind directions. The discrete wind directions were input into the NLR model using the usual one hot encoding, a.k.a. 1-of-$c$ coding scheme (Bishop, 1995; Hsieh, 2009), i.e. there were four binary indicator variables in the model input, where [1 0 0 0], [0 1 0 0], [0 0 1 0] and [0 0 0 1] represented the four wind directions. When extrapolating from the centre of the training data, NLR$_{OR}$ chose one of four possible centres, each computed using only the training data belonging to a particular wind direction.

Fig. 7a shows the difference in MAEn between NLR$_{OR}$ and NLR over outliers. That the mean and median positions of the waistline of the boxes over all the datasets as indicated by the horizontal dash and dot-dash lines, respectively, are below the horizontal axis indicates that NLR$_{OR}$ tends to do better than NLR on outliers. More importantly, there are many trials with values lying well below the lower whiskers of the boxes and none well above an upper whisker. Thus NLR$_{OR}$ has been effective in avoiding terrible extrapolation by NLR.

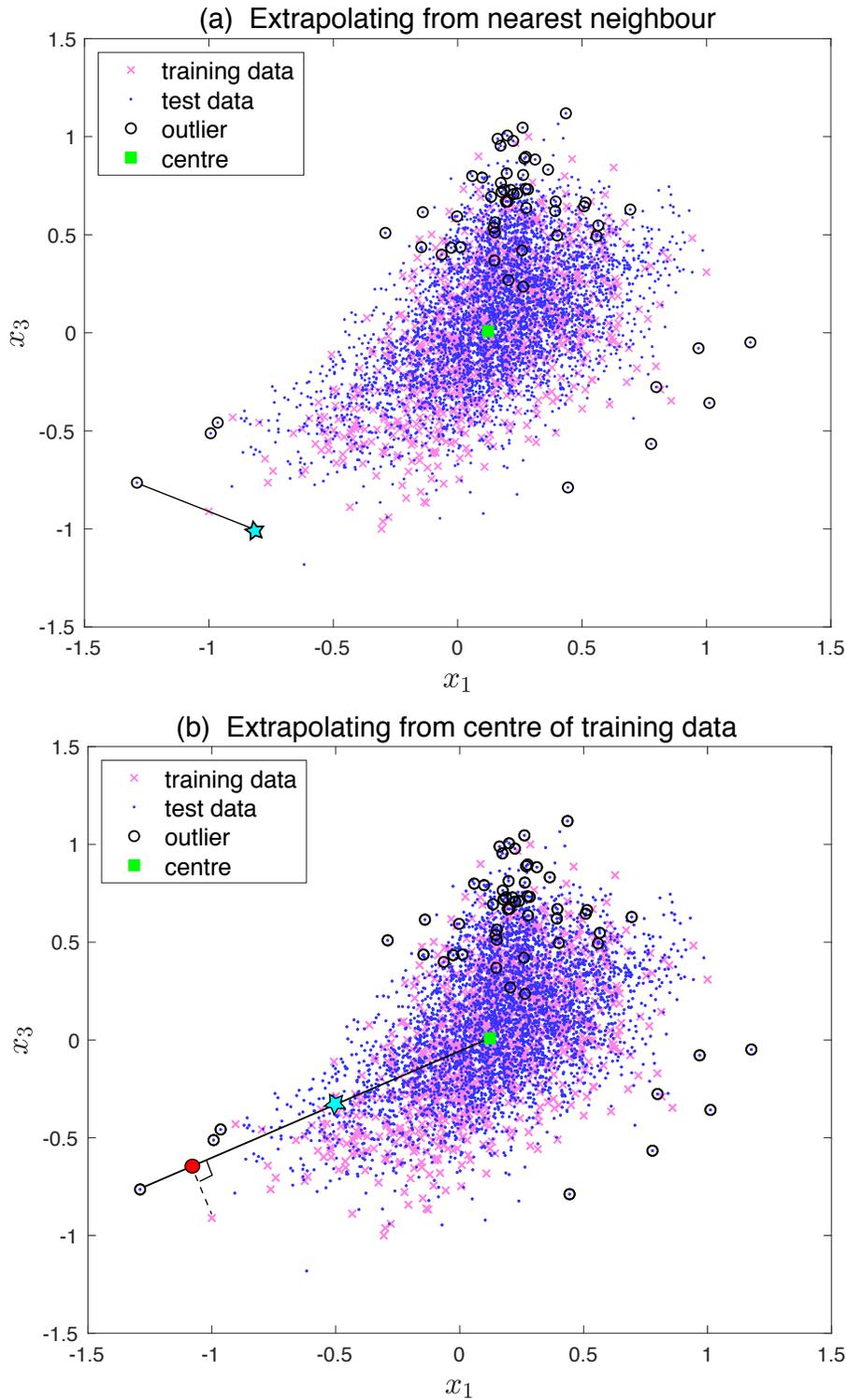

Figure 6: Linear extrapolation to outlying test data, (a) from the nearest neighbour and (b) from the direction of the centre of the training data, as seen in the $x_1$- $x_3$ space. Training data are indicated by crosses and test data by dots, with outliers circled. For the outlier to the left, linear extrapolation involves estimating the gradient of the NLR function from two points. In (a), the two points are the nearest neighbour and the point marked by a star and in (b) the solid circle (projection of the nearest neighbour onto the line joining the outlier and the centre) and the point marked by a star along that line.

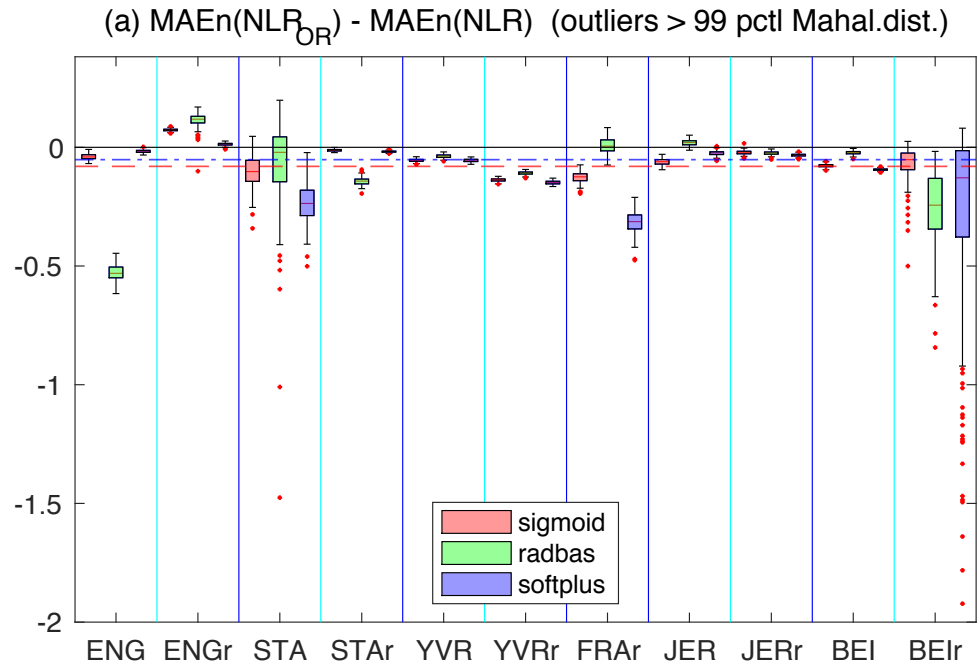

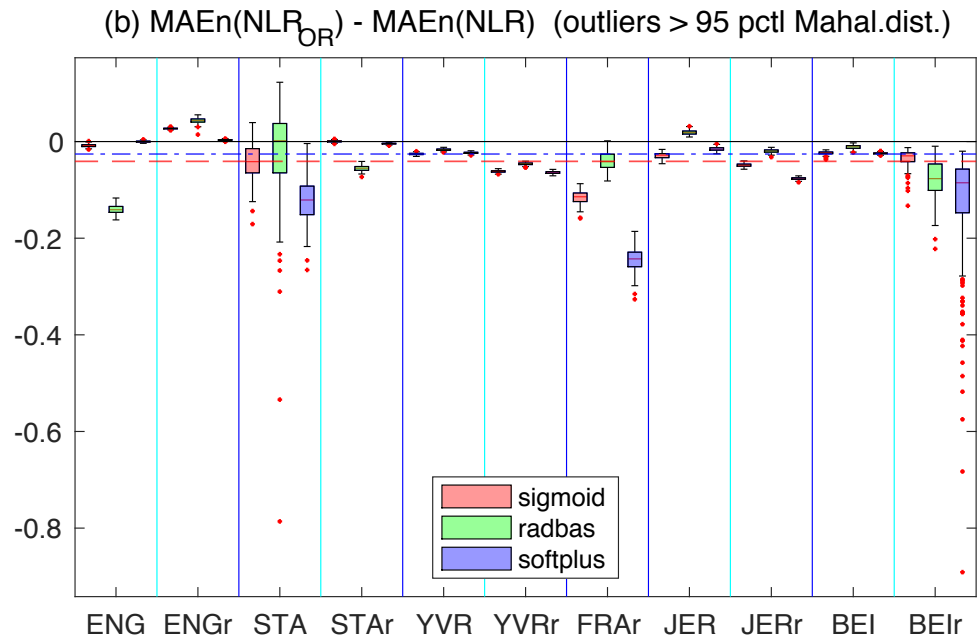

Figure 7: Boxplot of the difference in the normalized MAE between the $NLR_{OR}$ and NLR models for outliers defined as test data exceeding the (a) 99th and (b) 95th percentile of the Mahalanobis distance of the training data. The mean and the median position of the waistline over all the datasets are indicated by the horizontal dash and dot-dash lines, respectively. The rightmost box in (a) has three distant small crosses in the range $(-3.5, -2)$ which are not shown.

The choice of outliers, defined as input test data having a Mahalanobis distance (with respect to the training set) exceeding the 99th percentile of the Mahalanobis distance of the input training data, was somewhat arbitrary, so we also plotted out the result when the boundary for outliers was set at the 95th percentile (Fig. 7b), which would have more input points classified as outliers (Table 1). Fig. 7b generally resembles Fig. 7a, except for a different vertical scale. Similarly, the difference in the Spearman correlation between $NLR_{OR}$ and NLR over outliers (Fig. 8) confirmed the conclusions reached in Fig. 7.

Fig. 9 shows the difference in MAEn between $NLR_{OR}$ and LR over outliers. With the horizontal dash (mean) and dot-dash (median) lines both lying below the horizontal axis, $NLR_{OR}$ tended to do better than LR on outliers in MAEn. However, there are more trials with values lying above the upper whiskers of the boxes than below the lower whiskers. The general advantage of $NLR_{OR}$ over LR is more clear in Fig. 9b than in Fig. 9a, where outliers were defined as exceeding the 95th (instead of the 99th) percentile of the Mahalanobis distance of the training data.

For correlation (Fig.10), the comparison between $NLR_{OR}$ and LR is inconclusive, as the dot-dash line in both Figs. 10a and b and the dash line in Fig. 10b are almost indistinguishable from the horizontal axis. Together with the results from Figs. 4a and 5a, using LR to replace NLR predictions on outliers also appears to be effective in avoiding poor results from nonlinear extrapolation, though $NLR_{OR}$ tends to be better than LR (Fig. 9).

Next, we examine a particular dataset for bad nonlinear extrapolation behaviour. In Fig. 2a, BEIr, the Beijing $PM_{2.5}$ dataset with data from 2013-2015 used for training and 2010-2012 for testing, shows some trials lying well above the upper whiskers. For the NLR model, we chose the one using the sigmoidal activation function, which was the least ill-behaved among the three activation functions for BEIr. One input variable, the cumulated precipitation (Liang et al., 2016), reached 223.0 mm in the test data in July 2012, whereas the maximum value in the training data was 51.1 mm, i.e. this input in the test data was 4.36 times the maximum value in the training data, thereby forcing the NLR model to do an extreme extrapolation.

Each input variable $x_i$ ($i = 1,\ldots, d$) was nondimensionalized (i.e. normalized) based on its maximum and minimum values from the training dataset, i.e.

$$x_i' = 2\frac{x_i - x_{i\min}}{x_{i\max} - x_{i\min}} - 1, \tag{9}$$

before being fed into the NLR models, as recommended for the ELM Matlab code (http://www.ntu.edu.sg/home/egbhuang/elm_random_hidden_nodes.html). Thus the maximum and minimum values of $x_i'$ over the training data are 1 and −1, respectively. The maximum and minimum values were also found over the testing data for each variable $x_i'$, yielding $x_i'_{\max(test)}$ and $x_i'_{\min(test)}$. We define

$$r_{outl} = \max_i(|x_{i\max(test)}'|, |x_{i\min(test)}'|), \tag{10}$$

a useful measure of how bad the extrapolation situation is for the test dataset (after nondimensionalization). For BEIr, $r_{outl} = 7.73$ as listed in Table 1. This follows from the fact that the nondimensionalized cumulated precipitation in the test data reached a maximum of 7.73 (versus a maximum of 1.00 in the training data). Thus the ratio of $223/51.1 = 4.36$ in the raw cumulated precipitation translated into a much larger ratio of 7.73 in the nondimensionalized input.

To understand this effect, consider an input variable where the training dataset has the minimum value $a$ and the maximum $b$. Suppose the worst outlier test input has the value $c$, with $c > b$. Substituting these values into (9) and (10) yields

$$r_{outl} = \frac{2c - a - b}{b - a}. \tag{11}$$

For a variable bounded below by 0, one usually has $a \ll b$, hence

$$r_{outl} \approx \frac{2c}{b} - 1. \tag{12}$$

For large $c/b$, $r_{outl} \to 2c/b$, i.e. for an input variable bounded below by zero, $r_{outl}$ can be up to double the raw ratio $c/b$. For the strongest outlier in BEIr, with $c = 223$ mm, $b = 51.1$ mm and $c/b = 4.36$, (12) gives $r_{outl} = 7.73$, as indeed found in the nondimensionalized data.

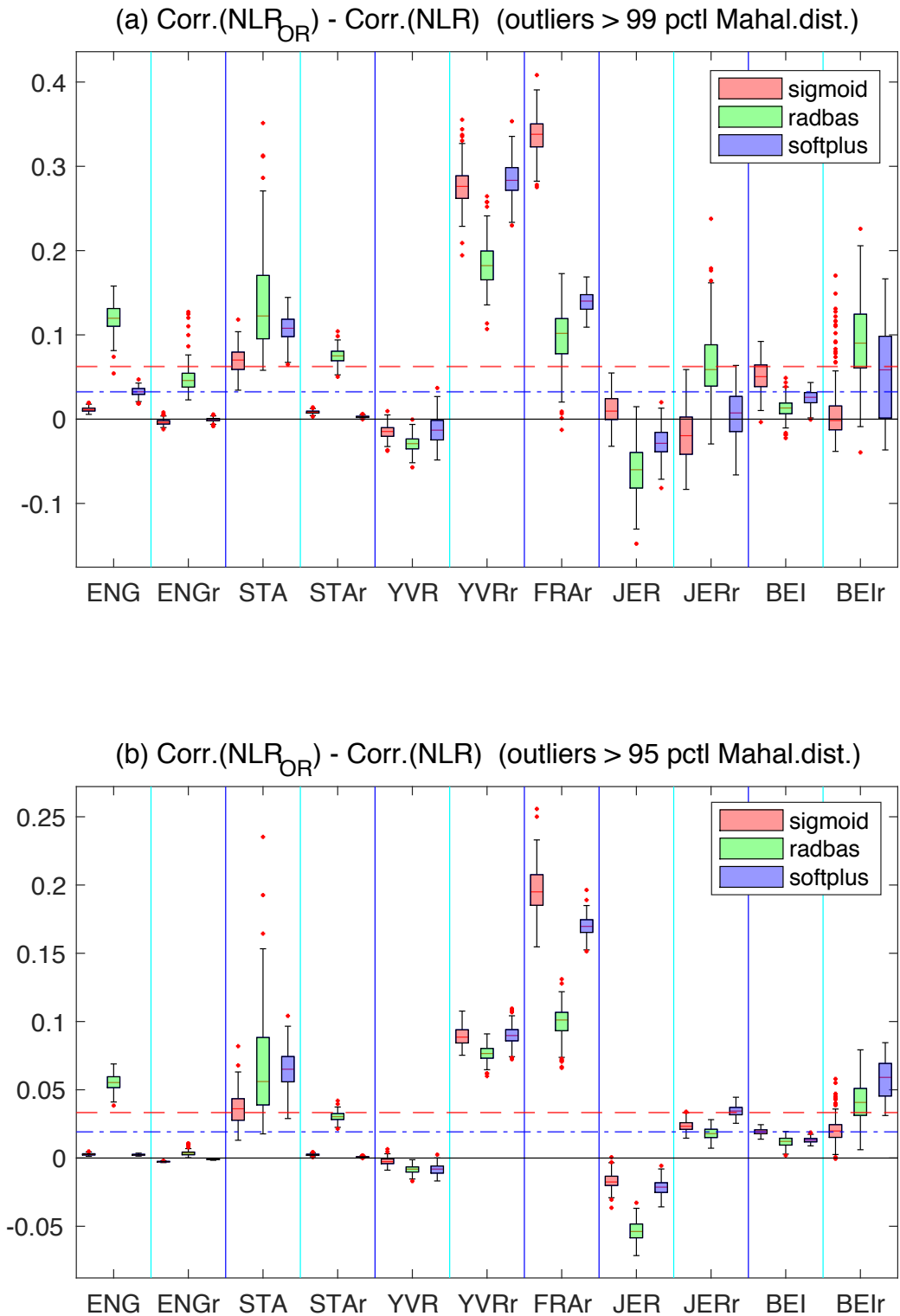

Figure 8: Boxplot of the difference in the Spearman correlation between the $NLR_{OR}$ and NLR models for outliers defined as test data exceeding the (a) 99th and (b) 95th percentile of the Mahalanobis distance of the training data.

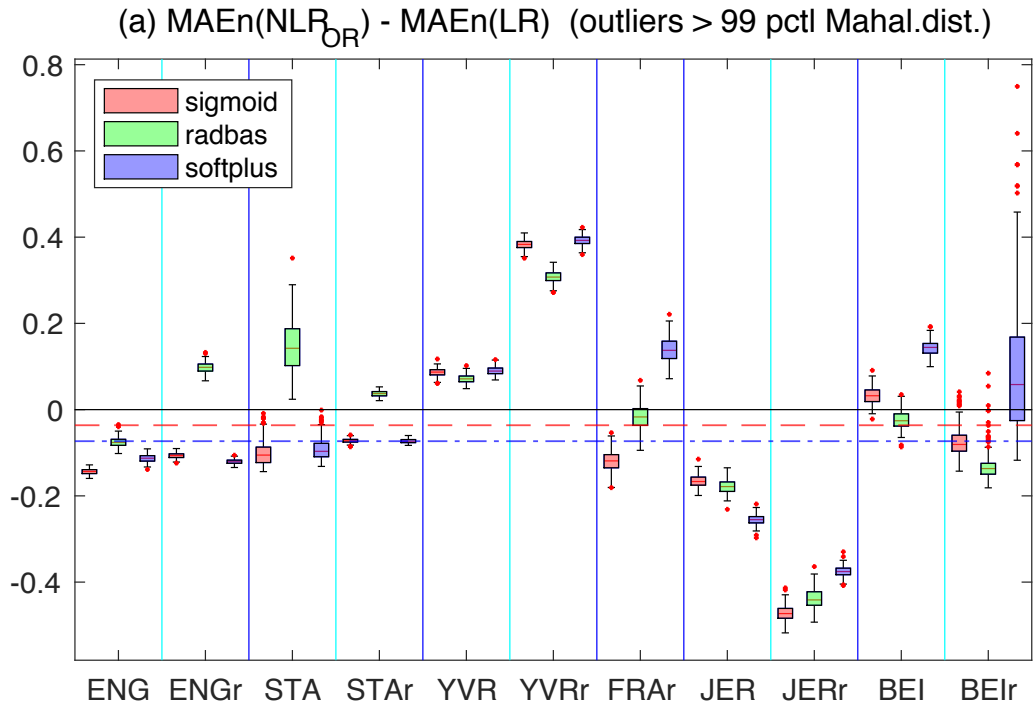

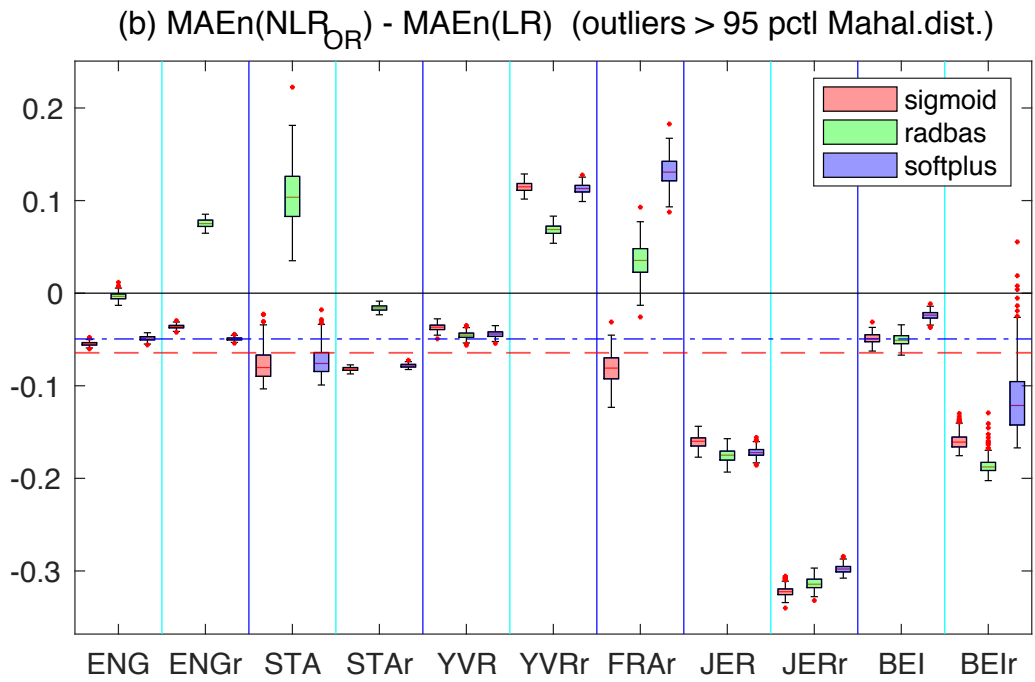

Figure 9: Boxplot of the difference in the normalized MAE between the $NLR_{OR}$ and LR models for outliers exceeding the (a) 99th and (b) 95th percentile of the Mahalanobis distance of the training data.

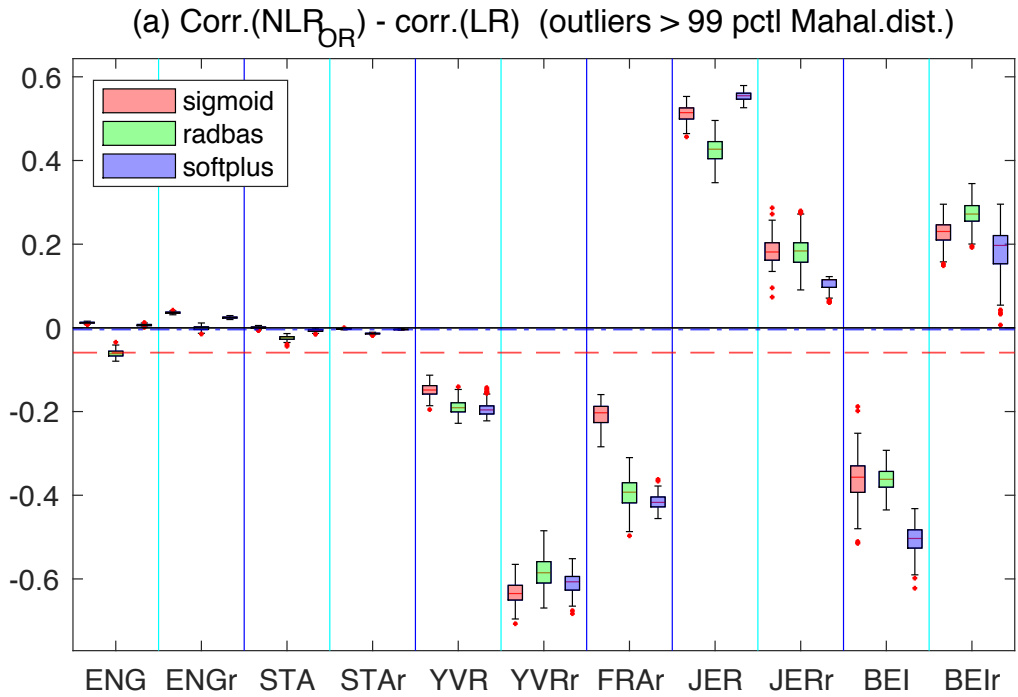

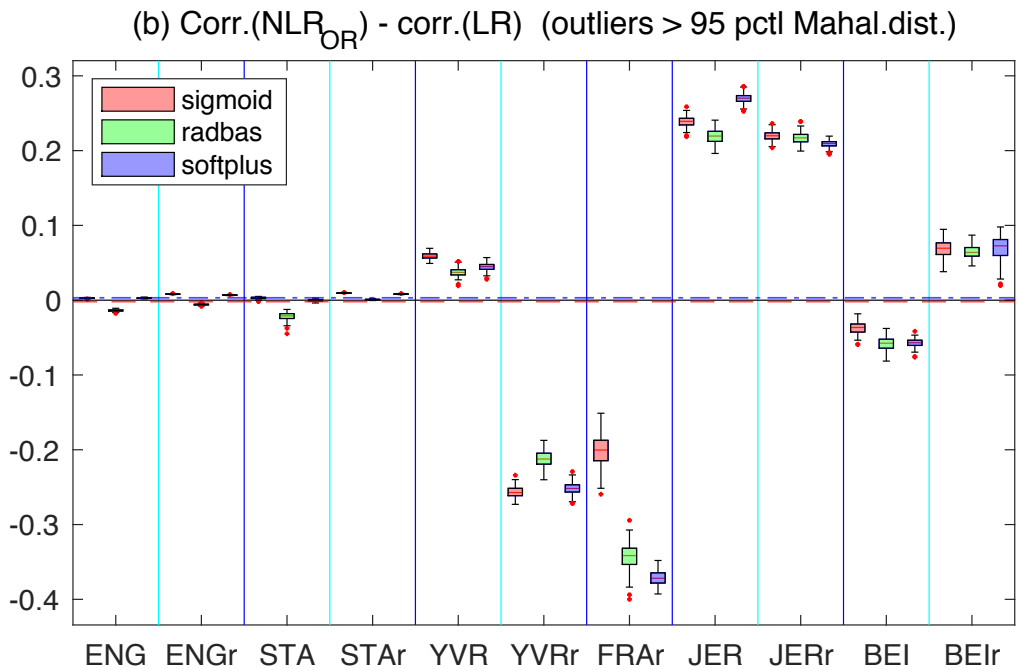

Figure 10: Boxplot of the difference in the Spearman correlation between the NLR$_{OR}$ and LR models for outliers exceeding the (a) 99th and (b) 95th percentile of the Mahalanobis distance of the training data.

A comparison with the toy problem of Fig. 1 gives some perspective on how extreme $r_{\text{outl}} = 7.73$ is. In the toy problem, the maximum test data point is $x = 6$, whereas the maximum training $|x|$ is about 2, hence $r_{\text{outl}} \approx 6/2 = 3$, i.e. the strong extrapolation illustrated in the toy problem is actually quite mild compared to extreme extrapolations encountered in real world problems.

Fig. 11a shows the observed $PM_{2.5}$ concentration over the 24 hours containing the extreme precipitation event, the output from 200 trials from the NLR and $NLR_{OR}$ models and from the LR model, while Fig. 11c shows the cumulative precipitation during the same 24 hours. In general, precipitation lowers the $PM_{2.5}$ concentration, and indeed during this intense precipitation event, the observed $PM_{2.5}$ concentration dropped to a low level. The high values predicted by some NLR trials during this precipitation events are puzzling (Fig. 11a). Since the observed $PM_{2.5}$ concentration is a non-negative quantity, when model predictions were negative, they were set to zero. In Fig. 11b, by not setting negative model predictions to zero, we found that for most trials, NLR (as well as LR and $NLR_{OR}$) gave negative predictions. By setting the negative predictions to zero, we were left with some trials showing high values in Fig. 11a, which contributed to Fig. 2a displaying some trials with MAEn lying well above the upper whiskers in the box for the sigmoid function run at BEIr. As each trial was obtained from the ensemble average of 100 members, it is worth examining the predictions of the individual ensemble members for one given trial. Fig. 11d shows most members giving large negative predictions, but also many members giving large positive predictions. Thus under extreme extrapolation, NLR ensemble members can not only predict an unrealistically large magnitude but also a wrong sign. While setting negative predicted values to zero is appropriate when the observed quantity is non-negative, it has the unintended bias of retaining trials with large positive predictions while eliminating those with negative predictions.

The second largest $r_{\text{outl}}$ in Table 1 occurred for STA, with $r_{\text{outl}} = 6.22$. On 9 March 2011, the input variable, water equivalent of accumulated snow depth, reached 231 mm whereas the maximum value in the training data was only 64 mm, i.e. the raw ratio $c/b = 3.61$. The estimate (12) gives $r_{\text{outl}} = 6.22$, as was indeed found from the nondimensionalized data. STA had the highest percentage of outliers beyond the 95th percentile to the total number of testing data in Table 1, with 23.4% instead of the typical value of around 5%. STA also displays the second most extensive spread in the results from 200 trials (Fig. 2a).

Fig. 12a shows NLR to severely overpredict the streamflow around days 110-135, while LR underpredicts. There is a complicated relation between accumulated snow depth and streamflow. When there is much accumulated snow, the water is stored in the snow and not released to the streamflow, so there is a negative correlation between the two, as manifested by LR underpredicting the streamflow when given extremely high snow depth. However, there is interaction between temperature and snow depth – if temperature is high, large snow depth means much snow melt to increase streamflow, whereas if temperature is low, there is no snow melt to feed the streamflow. $NLR_{OR}$ performed well during days 110-135 in both Fig. 12a showing the 200 trials and Fig. 12c showing the 100 ensemble members from a single trial.

In Table 1, the smallest $r_{\text{outl}}$ occurred for JER ($r_{\text{outl}} = 1.24$) and JERr ($r_{\text{outl}} = 1.26$), indicating that the outliers in the test input data were quite modest. Hence it is not surprising that the NLR performed well relative to LR for the outliers in the test data for these two (Fig. 4a). Also JER and JERr had the fewest outliers (24 and 17, respectively) exceeding the 99th percentile (Table 1), hence greater uncertainty than other datasets in estimating model performance over the outliers.

In short, $r_{\text{outl}}$ is a simple and useful indicator for how serious the outlier problem is for a particular dataset, with $r_{\text{outl}} > 1.5$ indicating the presence of strong outliers and $r_{\text{outl}} > 3$, extreme outliers, in the test input data. How much damage an outlier in a particular input variable can do depends on how strongly the NLR model weighs this input – i.e. an outlier in an unimportant predictor would do far less damage than a comparable outlier in an important predictor.

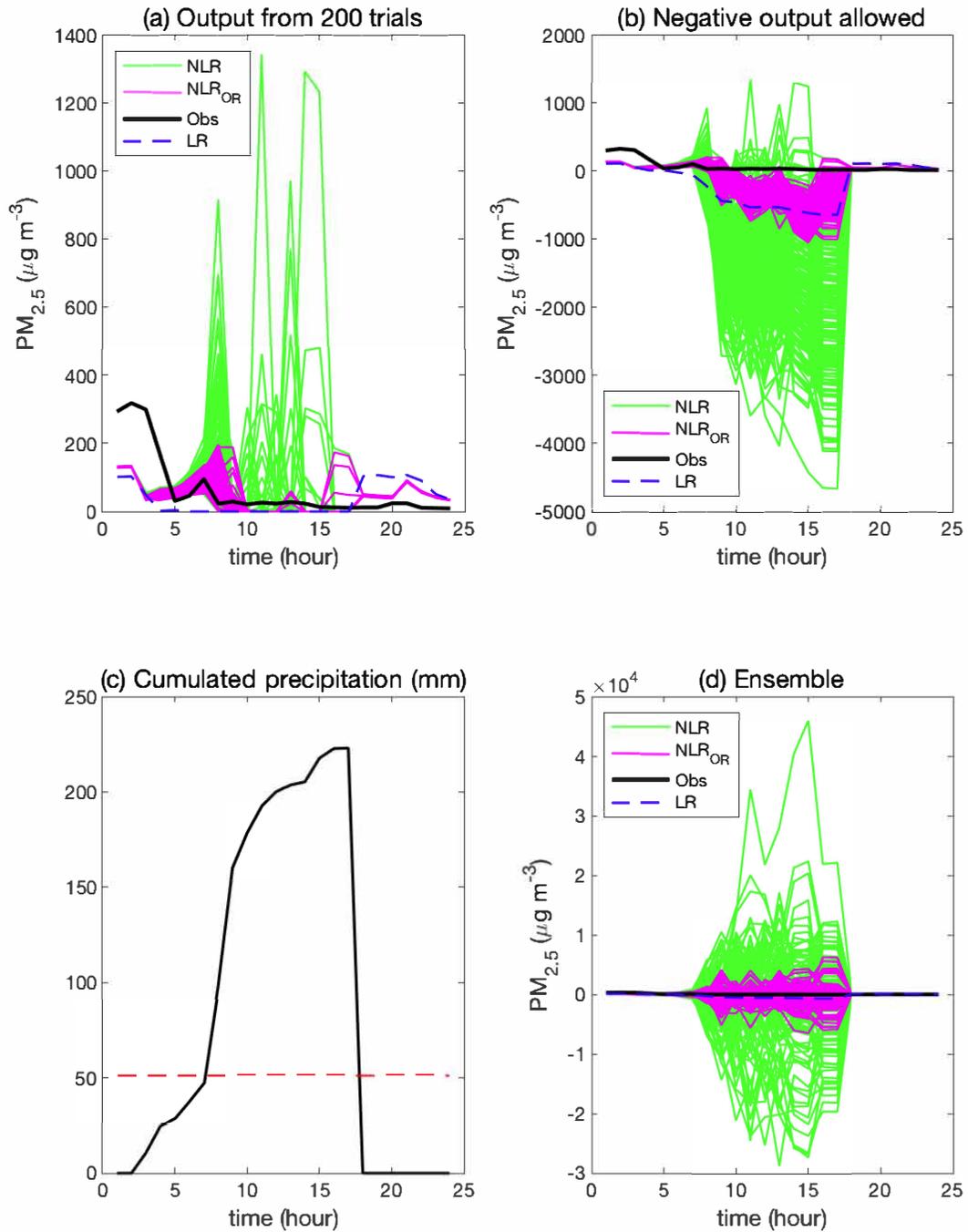

Figure 11: PM$_{2.5}$ at Beijing during the 24 hours in July 2012 containing the extreme precipitation event: (a) Output from 200 trials shown by thin lines for the NLR and NLR$_{OR}$ models, with observations and LR model output shown by the thick line and the dashed line, respectively. (b) Same as (a) but with model output allowed to be negative. (c) Cumulated precipitation, with dashed line indicating the maximum value encountered in the training data. (d) The 100 individual ensemble members from one trial. Note the different vertical scales used in (a), (b) and (d). The start time $t = 1$ hour in the plots corresponds to 21 July 2012, 11:00 Beijing time (UTC+8), with the PM$_{2.5}$ value being the average concentration between 11:00-12:00.

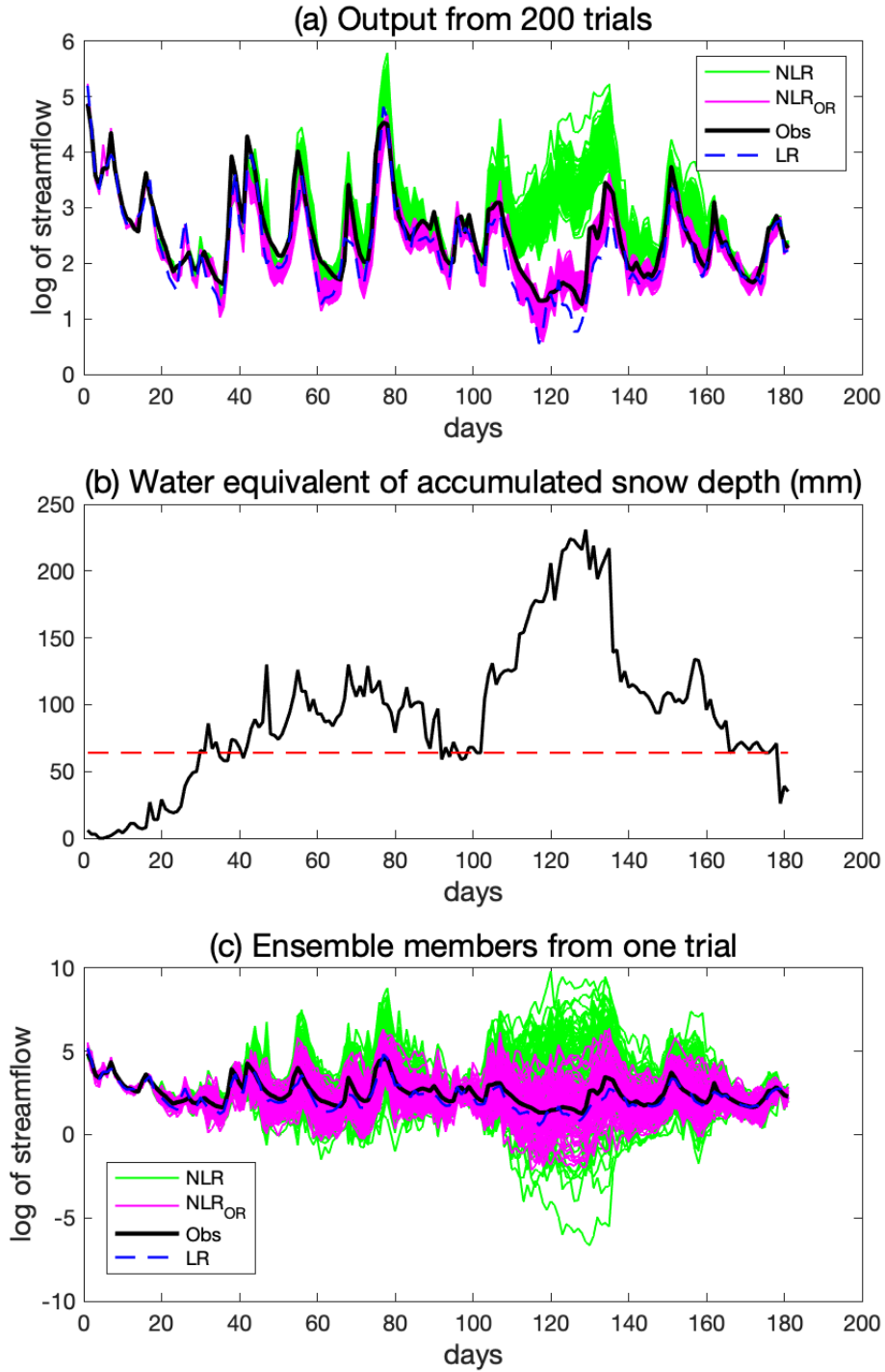

Figure 12: Streamflow (in $m^3\ s^{-1}$) at STA, transformed by the natural logarithm, during 1 Nov 2010 (day 1) to 30 Apr 2011 (day 181): (a) Output from 200 trials shown by thin lines for the NLR and $NLR_{OR}$ models, with observations and LR model output shown by the thick line and the dashed line, respectively. (b) Water equivalent of the accumulated snow depth, with dashed line indicating the maximum value encountered in the training data. (c) The 100 individual ensemble members from one trial. Note the different vertical scales used in (a) and (c).

# 6 Summary and Discussion

The relatively modest success of machine learning methods in environmental sciences when compared to the success in non-environmental fields is usually attributed to the fact that environmental datasets tend not to have a large number of independent observations, especially in climate datasets. This paper points to an overlooked factor, namely the rather different types of data used – mainly continuous input data in environmental sciences versus mainly discrete or categorical input data in non-environmental fields. Unlike discrete data, which usually have finite upper and lower bounds, and categorical data, continuous data in environmental sciences usually do not have a finite domain, so new input data can lie far outside the domain used in training the model. As nonlinear extrapolation is generally far more unpredictable than linear extrapolation, the suspicion is that nonlinear regression models may underperform simple LR on "outliers", i.e. test data where the input lies outside the domain of the training input data. The outliers could be caused by a combination of extreme events, bad data, climate change or regime shifts in the mean and/or the fluctuations, etc. Six environmental datasets were used to check the performance of NLR versus LR, revealing that the general outperformance of NLR over LR on non-outliers tended to turn into underperformance on outliers. Underperformance on outliers is particularly worrying since outliers are often associated with extreme environmental events, which have huge impact on human health, safety and economy. Furthermore, some environmental variables have heavy-tailed distributions (Cavanaugh et al., 2015), hence more common occurrence of outliers. Under climate change (Gaitan et al., 2014), changes or trends in weather and climate extremes (Heim, R. R., Jr., 2015) would further increase the number of test input data being outliers.

When new input data lie outside the domain of training input data, Occam's Razor would discard the complicated nonlinear model for a simpler model. The prediction scores over the six environmental datasets revealed that simply replacing NLR by LR for the outliers tended to improve performance. A more sophisticated alternative ($NLR_{OR}$) – linear extrapolation using the NLR model within the non-outlier domain – tended to outperform NLR, and to a lesser extend LR, for the outliers.

Although Occam's Razor is widely used in science, it is not infallible. There is no guarantee that our proposed use of linear extrapolation will outperform nonlinear extrapolation for any particular dataset. Another problem is that there is often not a unique way to implement Occam's Razor. For instance, in model selection, various information criteria have been developed to select the optimal model from a finite set of models (with different number of adjustable model parameters) so as to avoid overfitting the data with an excessively complex model. Akaike information criterion (AIC) (Akaike, 1974), Bayesian information criterion (BIC) (Schwarz 1978), etc. will generally select different models (von Storch and Zwiers, 1999, Sect. 12.2.11). Similarly, there can be alternatives to $NLR_{OR}$ for taming nonlinear extrapolation from input outliers.

The main weakness in the proposed Occam's Razor approach is the difficulty in defining an "outlier" precisely. To determine which test input data were "outliers", the Mahalanobis distance was computed for the training input data, and the 99th percentile of the Mahalanobis distance was considered the boundary for outliers. A test input data point having (i) a Mahalanobis distance exceeding the 99th percentile boundary and (ii) a (Euclidean) distance to the centre of the training data exceeding the distance between its nearest neighbour and the centre was considered an outlier. An arbitrariness lies in the choice of the percentile boundary (e.g. 98% or 95% could be chosen). Also the Mahalanobis distance may not be optimal for detecting outliers in non-Gaussian data (Chandola et al., 2009; Leys et al., 2018). Future work could involve finding better ways to identify the outliers.

The $NLR_{OR}$ approach can be applied to other continuous NLR models, e.g. back-propagating ANN (shallow and deep neural networks), support vector regression, Gaussian process, etc. In this paper, only the randomized neural network model (ELM) was used. The other models will perform nonlinear extrapolation somewhat differently, but the general notion that an NLR model usually behaves much more unexpectedly as it extrapolates outside the training domain than the LR model remains valid (Hsieh, 2009, pp.303-306).

The $NLR_{OR}$ approach would need some modification for non-continuous models like random forests (Breiman, 2001), which would give an inaccurate gradient for the linear extrapolation. A likely remedy would be to estimate the gradient many times using different points in the non-outlier domain and average the multiple estimates.

In conclusion, it is generally assumed that an ML model which has been properly regularized to avoid overfitting to noise in the data (e.g. by cross validation) (Bishop, 2006) will perform well with new data. What this paper has shown is that merely focusing on not overfitting/underfitting is not enough for environmental scientists/managers using unbounded continuous input data – where new input can easily lie well outside the domain of the input data used for training the model, with nonlinear extrapolation having the potential to inflict terrible damage to model performance. Hence, the following procedure is recommended:

(a) New input data are screened for outliers relative to the training input data (using e.g. the simple Mahalanobis distance based scheme in this paper).

(b) For strong or extreme outliers, avoid the nonlinear extrapolations from the NLR model. Instead, either use LR or $NLR_{OR}$, or issue a warning that no reliable prediction can be made (as LR and $NLR_{OR}$ still involve linear extrapolation).

Similarly, when environmental scientists write papers comparing NLR models with LR, it is not reasonable to include the input outliers in the test/validation datasets when computing the performance scores (e.g. MAE, RMSE, correlation, etc.). Only by testing/validating NLR within the domain it has been trained on can there be a fair comparison between NLR and LR. Only then, can the haunting conundrum that ML models used for NLR do not perform much better than traditional LR models be finally vanquished, and ML models be embraced by environmental scientists/managers.

## Acknowledgements

The author is grateful for the assistance received in obtaining the six datasets – Dr. Aranildo R. Lima for the four datasets from British Columbia, Dr. Yiwen Mao for the Jersey wind dataset and Dr. Xuan Liang for the Beijing $PM_{2.5}$ dataset. The Matlab code for the ELM model was downloaded from `http://www.ntu.edu.sg/home/egbhuang/elm_random_hidden_nodes.html`. Dr. Benyang Tang gave helpful comments on an earlier version of the manuscript.

Lima, A. R., Cannon, A. J., and Hsieh, W. W. (2016). Forecasting daily streamflow using online sequential extreme learning machines. *J. Hydrology,* 537, 431–443.

Lima, A. R., Hsieh, W. W., and Cannon, A. J. (2017). Variable complexity online sequential extreme learning machine, with applications to streamflow prediction. *J. Hydrology,* 555, 983–994.

Mahalanobis, P. C. (1936). On the generalised distance in statistics. *Proceedings of the National Institute of Sciences of India,* 2(1), 49–55.

Mao, Y. and Monahan, A. (2018). Linear and nonlinear regression prediction of surface wind components. *Clim. Dynam.*, 51, 3291-3309 .

Peng, H., Lima, A. R., Teakles, A., Jin, J., Cannon, A. J., and Hsieh, W. W. (2017). Evaluating hourly air quality forecasting in Canada with nonlinear updatable machine learning methods. *Air Quality, Atmosphere & Health,* 10, 195–211.

Rasouli, K., Hsieh, W. W., and Cannon, A. J. (2012). Daily streamflow forecasting by machine learning methods with weather and climate inputs. *J. Hydrology,* 414, 284–293.

Schmidt, W. F., Kraaijveld, M. A., and Duin, R. P. W. (1992). Feed forward neural networks with random weights. In *11th IAPR International Conference on Pattern Recognition, Proceedings,* Vol II: Conference B: Pattern Recognition Methodology and Systems, pages 1–4. Int. Assoc. Pattern Recognition.

Schwarz, G. (1978). Estimating the dimension of a model. *Annals of Statistics,* 6, 461-464.

Spearman, C. (1904). The proof and measurement of association between two things. *American Journal of Psychology,* 15(1), 72–101.

von Storch, H. and Zwiers, F.W. (1999). *Statistical Analysis in Climate Research.* Cambridge Univ. Pr.

Vapnik, V., Golowich, S. E., and Smola, A. (1997). Support vector method for function approximation, regression estimation, and signal processing. In Mozer, M. C., Jordan, M. I., and Petsche, T., editors, *Advances in Neural Information Processing Systems,* 9, 281–287.

Yuval and Hsieh, W. W. (2002). The impact of time-averaging on the detectability of nonlinear empirical relations. *Quarterly Journal of the Royal Meteorological Society,* 128, 1609–1622.

Zeng, Z., Hsieh, W. W., Burrows, W. R., Giles, A., and Shabbar, A. (2011a). Surface Wind Speed Prediction in the Canadian Arctic using Non-Linear Machine Learning Methods. *Atmos.-Ocean,* 49(1), 22–31.

Zeng, Z., Hsieh, W. W., Shabbar, A., and Burrows, W. R. (2011b). Seasonal prediction of winter extreme precipitation over Canada by support vector regression. *Hydrology and Earth System Sciences,* 15(1), 65–74.
## A  Datasets

• ENG and STA

Daily streamflow (1985–2011) for ENG, the Englishman River station (49° 19′ 0″ N, 124° 16′ 58″ W) and STA, the Stave River station (49° 33′ 21″ N, 122° 19′ 24″ W), (obtained from Water Survey of Canada (http://wateroffice.ec.gc.ca/), have been studied previously (Rasouli et al., 2012; Fleming et al., 2015; Lima et al., 2015). Using the same data as in Lima et al. (2016); Lima et al. (2017) (which gave more details on the data source), streamflow at ENG and STA were predicted 1 day in advance ($t$ +1) using local hydro-meteorological observations and weather forecasts. This study used 34 predictors: (i) Local observations including streamflow, precipitation, maximum temperature and minimum temperature at day $t$, $t-1$ and $t-2$, and 7-day moving averages of precipitation, maximum temperature, and minimum temperature, and (ii) weather forecasts at day $t$ +1, from the Global Ensemble Forecast System (GEFS) Reforecast dataset (Hamill et al., 2013), including mean sea level pressure, precipitable water, 2-m specific humidity, 2-m temperature, volumetric soil moisture content, soil temperature, water runoff, water equivalent of accumulated snow depth, wind speed, all of them at 0900 and at 2100 UTC, plus the total 24-h precipitation. Only days not containing missing values were used (9540 days for ENG and 9620 days for STA). As streamflow has a right-skewed distribution, the natural logarithm transformation was applied first to the streamflow data prior to performing the regression (Lima et al., 2017). Table 1 gives additional information on the datasets.

- YVR

Daily precipitation (1971-2000) from the Vancouver International Airport YVR (WMO station 71892, 49° 12.0′ N, 123° 10.8′ W) was used for statistical downscaling, i.e. coarse resolution general circulation model (GCM) outputs were used to predict precipitation at local scale (Cannon, 2011; Lima et al., 2015). Three predictors, namely the sea-level pressure, 700-hPa specific humidity, and 500-hPa geopotential height, from the nearest National Centers for Environmental Prediction/National Center for Atmospheric Research (NCEP/NCAR) Reanalysis (Kalnay et al., 1996) grid point at 50°N, 122.5°W were used to predict the precipitation amount on wet days (defined as days with precipitation > 0.001 mm). As the precipitation amount was not normally distributed, the variable was transformed by taking the fourth root prior to performing regression, as commonly done for the precipitation amount (Khan et al., 2006).

- FRA

The FRA (Fraser River at Hope station in British Columbia, Canada) dataset contains daily observations of suspended sediment concentration (SSC) ($mg\,L^{-1}$) and streamflow ($Q$) ($m^3\,s^{-1}$) for the years 1970-1979 (Cannon, 2012; Lima et al., 2015). To reduce the non-Gaussian nature of the distribution, the $\log_{10}$ transformation was applied to the output variable (SSC at zero lead time) and the four predictors were logQ, dQ5, dQ30, and dQ90 (the 5-, 30-, and 90-day moving averages of the daily changes in $Q$). In Table 1, FRAr was shown but FRA (with 1970-1972 used for training and 1973-1979 for testing) was omitted since it only had 3 outliers in the test data (with outliers defined as exceeding the 99th percentile of the Mahalanobis distance).

- JER

The daily winter (DJF) surface wind speed at Jersey Airport (JER, 49.2096°N, 2.1943°W, WMO station 03895) during 1980-2012 was predicted (at zero lead time) using four predictors at the mid-tropospheric level, namely temperature, geopotential height, zonal and meridional wind components at 500 hPa (Mao and Monahan, 2018), from the NCEP-Reanalysis 2 (Kanamitsu et al., 2002) 2.5° × 2.5° grid overlying the airport. As wind speed is non-negative, any negative predicted value would automatically be set to zero.

- BEI

A common air quality measure of fine inhalable particles with diameters ≤ 2.5 μm is the $PM_{2.5}$ concentration. Hourly $PM_{2.5}$ concentration measured by the US Embassy in Beijing (BEI) (Liang et al., 2015, 2016) was predicted (at zero lead time) by meteorological data from the Central Meteorological Agency of China during 2010-2015, with the data downloaded from the UCI Machine Learning Depository site (`https://archive.ics.uci.edu/ml/datasets/PM2.5+Data+of+Five+Chinese+Cities`). We used five continuous predictors (dew point, pressure, temperature, cumulative precipitation and cumulative wind speed) and four binary predictors indicating one of four general wind directions (NW, NE, S or CV (calm and variable)). As $PM_{2.5}$ concentration is non-negative, any negative predicted value would automatically be set to zero.